# Survival Prediction of Children Undergoing Hematopoietic Stem Cell Transplantation Using Different Machine Learning Classifiers by Performing Chi-squared Test and Hyper-parameter Optimization: A Retrospective Analysis

Ishrak Jahan Ratul[a, c], Ummay Habiba Wani[a, c], Mirza Muntasir Nishat[a, c], Abdullah Al-Monsur[a, c], Abrar Mohammad Ar-Rafi[a, c], Fahim Faisal[a, c], and Mohammad Ridwan Kabir[b, c]

[a] *Department of Electronic and Electronic Engineering*
[b] *Department of Computer Science and Engineering*
[c] *Islamic University of Technology (IUT), Gazipur, Bangladesh.*
{ishrakjahan, ummayhabiba, mirzamuntasir, al-monsur, abrarmohammad, faisaleee, ridwankabir}@iut-dhaka.edu

Bone Marrow Transplant (BMT), a gradational rescue for a wide range of neoplastic, chronic, and allied disorders emanating from the bone marrow, is an efficacious surgical treatment. Several risk factors, such as post-transplant illnesses, new malignancies, and even organ damage, can impair long-term survival after BMT. Therefore, technologies like Machine Learning (ML) are being linked to hematology for investigating the survival prediction of BMT receivers along with the influences that limit their resilience. In this study, a comprehensive approach is undertaken where an efficient survival classification model is presented, incorporating the Chi-squared feature selection method to address the dimensionality problem and Hyper Parameter Optimization (HPO) to increase accuracy. A synthetic dataset is generated by imputing the missing values, transforming the data using dummy variable encoding, and compressing the dataset from 59 features to the 11 most correlated features using Chi-squared feature selection. The dataset was split into two sections (train and test set) with a ratio of 80:20, and the hyper-parameters were optimized using Grid Search Cross-Validation (GSCV). Several supervised ML methods were trained in this regard, like Decision Tree (DT), Random Forest (RF), Logistic Regression (LR), K-Nearest Neighbors (KNN), Gradient Boosting Classifier (GBC), Ada Boost (AdB), and XG Boost (XGB). The simulations have been performed for both the default and optimized hyperparameters by using the original and reduced synthetic dataset. After ranking the features using the Chi-squared test, it was observed that the top 11 features with HPO, resulted in the same accuracy of prediction (94.73%) as the entire dataset with default parameters. Moreover, this approach requires less time and resources for predicting the survivability of children undergoing BMT. Hence, the proposed approach may aid in the development of a computer-aided diagnostic system with satisfactory accuracy and minimal computation time by utilizing medical data records.

## 1 INTRODUCTION:

Even in its most curable forms, cancer kills millions of people every year. According to the cancer statistics of 2020, the estimated death tolls in the USA from colon, pancreatic, lung, breast, and prostate cancers are 53200, 47050, 135720, 42690, and 33330, respectively [1]. When there is no cure, physicians endeavor to extend the lifespan of a cancer patient through surgery, radiation therapy, or chemotherapy as alternative methods of cancer treatment [2]. For various reasons, the high dose of medication during chemotherapy or radiation therapy causes bone marrow damage in patients [2]. Bone Marrow (BM), the delicate, elastic, adipose tissue located inside most skeleton structures, is responsible for creating the red blood cells of human blood [3-4]. It also contains Hematopoietic Stem Cells (HSC) that are merely immature blood-forming stem cells that are endowed with idiosyncratic properties like self-renewal, and they form populations of progenitor cells through cell division and differentiation [4-6]. However, the concept of BMT, otherwise known as Hematopoietic Stem Cell Transplant (HSCT), gleans from the postulation of eliminating dysfunctional body parts and replacing them with healthy ones [7]. Although it is a lifesaving treatment, it has potential life-threatening risks [8]. Clinical HSCT commenced in 1957, at a time when the health domain was inadequately fathomed about HSCs, immunological reactions to transplants, and even about the specification of the antigens steering the course of action [9]. HSCT is not a surgery, but rather a specialized treatment for people afflicted by specific cancers or certain medical conditions [10-11]. The target of such a therapy is transfusing functional BM into a patient, subsequent to their own diseased BM being medicated for exterminating the aberrant cells [11]. The three prime objectives of HSCT are – (a) replacement of deceased stem cells affected by chemotherapy, (b) replacement of diseased marrow that is impotent to synthesize its endemic progenitor cells, and (c) infusion of allografts to assist in locating and destroying malignant cells [12].

Healthy BM can be either extracted from the patient (autologous transplant) or conferred on by a volunteer donor (allogeneic transplant). In the case of autologous transplant, stem cells come from any other healthy organ of the patient [3]. And for an allogeneic transplant, a donor with closely matched Human Leukocyte Antigens (HLAs) is needed [10]. Most of the times siblings having the same parents make for the closest matches, although other close relatives or perhaps an unrelated patron can also be a successful match. There are two ways of collecting donor stem cells for transplant – (a) BM collection, and (b) leukapheresis [13]. When a patient receives highly matched proteins from a donor, the odds of developing a severe adverse reaction, known as Graft-Versus-Host Disease (GVHD), are minimized [3]. Given that a donor cannot be found, cord blood transplants (stem cells collected from the umbilical cord), parent-child, and HLA haplotype mismatched transplants (stem cells collected from a parent, child, or sibling) can be performed [3]. HSCT is broadly adopted for hematopoietic system-acquired and congenital illnesses. According to the Health Resources & Services Administration, almost 23 million people have registered on the donor registry. Besides, the donor registry currently contains approximately 305,000 units of cord blood. The National Cord Blood Inventory (NCBI) provides around 112,000 units, which is reflected in this number, with an additional 4,000 units projected to be available in 2020. The Center for International Blood and Marrow Transplant Research (CIBMTR) registered a total of 9,267 related and unrelated BMTs conducted in the United States in 2018 [14]. According to a survey, undertaken by UPMC Children's Hospital of Pittsburgh, the percentages of patients who survived 100 or more days well after the transplant procedure, the percentage of patients who died of causes other than the underlying disease, and the percentage of patients who survived one or more years after the transplant procedure are 100%, 3%, and 94%, respectively [15].

To summarize, BMT is a treatment that saves and risks life at the same time. Hence, a lot of data collection and generation are required before the therapy. Classification techniques based on ML can be beneficial for disease prediction in a variety of healthcare situations. It has significant predictive ability for this type of problem and has been extensively used in recent years in a variety of sophisticated healthcare systems [16-20]. Moreover, it has lately been shown to be incredibly effective in the healthcare arena [21-25]. In this study, the survival prediction of children who received BMT was thoroughly investigated using seven supervised ML classifiers, such as: Decision Tree (DT), Random Forest (RF), Logistic Regression (LR), K-Nearest Neighbors (KNN), Gradient Boosting Classifier (GBC), Ada Boost (AdB), XG Boost (XGB), and a dataset obtained from the UCI ML repository [26]. The Chi-squared feature ranking technique was deployed after preprocessing the dataset to discover the important factors of survivability [27]. The entire study consists of four experiments, such as – (A) with a full set of features and default hyper-parameters, (B) with a full set of features and Hyper Parameter Optimization (HPO), (C) with a reduced feature dataset (based on the Chi-squared Test) and default hyper-parameters, and (D) with a reduced feature dataset (based on the Chi-squared Test) and Hyper Parameter Optimization (HPO) followed by a rigorous quantitative and qualitative analysis. An overall workflow diagram of this work is depicted in Figure 1.

The contributions of this study may be summarized as follows – (1) development of a suitable predictive model from raw data, (2) determination of critical factors influencing post-BMT survival, and (3) improvement of the prediction accuracy by reducing dimensionality problems. To the best of our knowledge, this dataset has never been exploited and analyzed in this way before, and it may appear as a significant contribution in helping the healthcare industry to develop a more trustworthy e-healthcare system and create a new horizon in the medical sector.

The following sections of the work include an extensive literature review followed by methodology, where dataset description, chi-squared test, hyperparameter optimization, and overall workflow diagram are depicted comprehensively. Hence, the simulation results are portrayed for four experimental setups, and corresponding discussions are presented in an elaborate fashion. Lastly, the conclusive remarks are carried out.

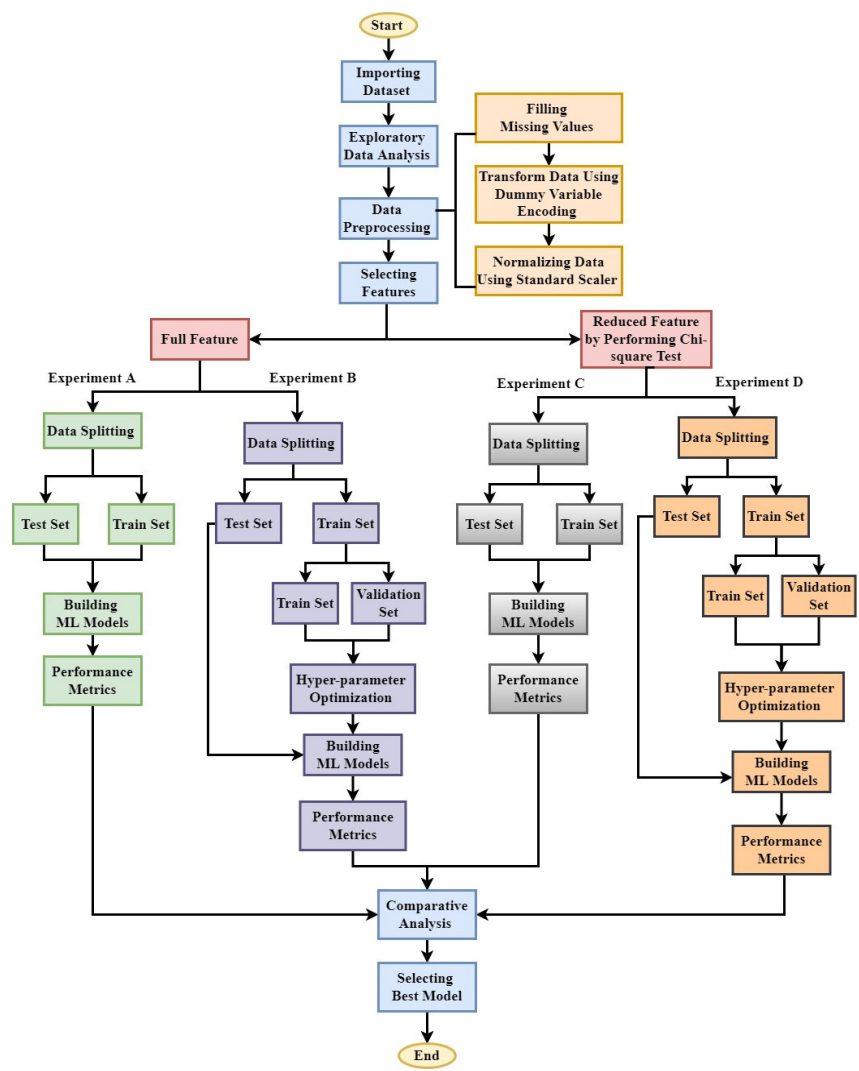

Figure 1: Overall flow diagram.

## 2 LITERATURE REVIEW

In recent times, ML techniques have been extensively exploited for diagnosis, prognosis, and therapeutics in the healthcare sector. Its applications are not limited to treatment procedures; rather, they are expeditiously gaining traction in a variety of research fields. In a narrative review, Nathan et al. highlighted essential ML concepts for novice readers, discussed the applicability of ML in hematology-related malignancies, and indicated key points for practitioners to consider before evaluating ML studies [28]. Vibhuti et al. also conducted a comparative evaluation of ML methods utilized in the discipline of HSCT, examining the categories of dataflows incorporated, designated ML algorithms used, and therapeutic consequences monitored [29]. On the other hand, patients with Acute Leukemia (AL) undergoing HSCT from unrelated donors exhibit a plethora of variations, even after rigorous genetic matching. To address this, Ljubomir et al. sought to develop an algorithm to predict the five-year survival of AL patients post-allogeneic transplant [30]. Similarly, Brent et al. trained a Bayesian ML model to predict acute GVHD, including mortality by day 180 [31]. However, with better donor data collection, it is possible to generate a more precise approximation of individual donor availability, as estimating group averages to the distinct donor is an untrustworthy proposition. As a solution to this problem, Adarsh et al. suggested an ML-based technique for estimating the availability of each listed donor and validation of forecasting accuracy [32]. Additionally, Ying Li et al. focused on creating and verifying an ML technique for estimating donor availability, implementing and comparing three ML algorithms [33]. As a result of organized registry establishments and biological data incorporation, data procured from HSCT institutions is becoming highly proliferated and labyrinthine. Consequently, conventional statistical methods are confirmed to be obsolete. In its provision, Shouval et al. aimed to advocate the implementation of ML and Data Mining (DM) schemes in the study of HSCT, covering transplant performance prognosis as well as donor selection [34]. Similarly, Jan-Niklas et al. explored current ML breakthroughs in the Acute Myeloid Leukemia (AML) diagnosis as a prototype condition encompassing hematologic neoplasms [35]. Further, the goal of the research by Liyan et al. was to shape an ALL (Acute Lymphocytic Leukemia)-relapse detection scheme relying on ML methods [36]. In addition, using Alternating Data Tree (ADTree), Kyoko et al. endeavored to design a model for predicting leukemia recidivism within a year following transplantation [37]. For contemplative and prospective analysis, ADTree was also employed by Yasuyuki et al. to scan databases containing information about adult patients with HSCT in Japan [38]. Daniela et al. also examined the organic phenomena associated with self-regeneration and augmentation of Hormone-Sensitive Prostate Cancer (known as CD34+ cells) in stable conditions and subsequent transplantation [39]. Moreover, a DM analysis involving 28,236 registered adult HSCT receiving patients from the European Group for Blood and Marrow Transplantation's AL registry was done by Shouval et al. to predict 100-day overall and non-relapse mortality, free of leukemia, and 2-year overall survival. The ADTree algorithm was employed to create models using 70% of the data set, and the remaining 30% of the data was utilized to validate them [40]. Moreover, in [41], a total of 25,923 adult AL cases were studied from the EBMT registry, using an in-silico approach.

In this study, an ML stratagem was adopted for eliciting a prediction of the survival rate of patients who had BMT or HSCT. All the previous works augmented the prediction study and related investigation through distinctive strategies; however, all have limitations that need to be overcome. The sole purpose of this research is to investigate whether HPO along with a reduced feature set can provide a reliable outcome using an investigative ML approach and to distinguish the most impactful factors on the children's survival who have received BMTs.

## 3 METHODOLOGY:

### 3.1 Dataset Description:

The dataset used in this study, was retrieved from the ML repository at the University of California, Irvine, and the version utilized in this study was extracted from [42]. It covers medical information for children who have been diagnosed with a variety of hematologic diseases and who underwent unmodified allogeneic unrelated donor HSCT. Hence, this dataset comprises 187 occurrences and 37 attributes that contain information about individuals who have been diagnosed with a range of hematologic malignant or benign diseases [26]. Most of the properties contain categorical data, while others contain Boolean and numerical values. The dataset's attributes are listed in Table 1. Following data extraction, it was subjected to exploratory data analysis using *Jupyter Notebook* and *Python* to find out the dataset's properties. The dataset has many categorical attributes and missing values. The statistical data for the dataset's numerical variables are summarized in Table 2.

Table 1: Dataset attributes.

| Sl. No. | Attribute | Type | Information |
|---|---|---|---|
| 1. | donor_age | Numeric | The donor's age at hematopoietic stem cell apheresis |
| 2. | donor_age_below_35 | Boolean | Is the donor under the age of 35? |
| 3. | donor_ABO | Categorical | The hematopoietic stem cell donor's ABO blood group |
| 4. | donor_CMV | Categorical | Cytomegalovirus infection prior to transplantation in the donor of hematopoietic stem cells |
| 5. | recipient_age | Numeric | The recipient's age at hematopoietic stem cell apheresis |
| 6. | recipient_age_below_10 | Boolean | Is the recipient's age under ten? |
| 7. | recipient_age_int | Categorical | Distinct intervals of the recipient's age |
| 8. | recipient_gender | Categorical | The recipient's gender |
| 9. | recipient_body_mass | Numeric | Mass of the transplanted hematopoietic stem cell recipient |
| 10. | recipient_ABO | Categorical | The recipient's ABO blood group |
| 11. | recipient_rh | Categorical | The Rh factor is present on the recipient's red blood cells |
| 12. | recipient_CMV | Categorical | Cytomegalovirus infection prior to transplantation in the donor of hematopoietic stem cells |
| 13. | disease | Categorical | Disease classification |
| 14. | disease_group | Categorical | Malignant or nonmalignant |
| 15. | gender_match | Categorical | Gender compatibility between donor and recipient |
| 16. | ABO_match | Categorical | HSC donor-recipient blood group compatibility |
| 17. | CMV_status | Categorical | Serological compatibility of hematopoietic stem cell donors and recipients based on CMV infection prior to transplantation |
| 18. | HLA_match | Categorical | Antigen compatibility between the donor and receiver of hematopoietic stem cells |
| 19. | HLA_mismatch | Categorical | HLA mismatches or matches |
| 20. | antigen | Categorical | How many antigens differ between the donor and receiver |
| 21. | allel | Categorical | How many alleles differ between the donor and receiver |
| 22. | HLA_group_1 | Categorical | The donor-recipient difference |
| 23. | risk_group | Categorical | Group at risk |
| 24. | stem_cell_source | Categorical | Hematopoietic stem cell source |
| 25. | tx_post_relapse | Boolean | The second bone marrow transplant following recurrence |
| 26. | CD34_x1e6_per_kg | Numeric | CD34kgx10d6 - CD34+ dose of cells per kg mass of recipient |
| 27. | CD3_x1e8_per_kg | Numeric | Dose of CD3+ cells per kilogram of recipient weight |
| 28. | CD3_to_CD34_ratio | Numeric | The ratio of CD3+ cells to CD34+ cells |
| 29. | ANC_recovery | Numeric | Time required for neutrophil recovery is defined as a neutrophil count greater than 0.5 x 10^9/L |
| 30. | PLT_recovery | Numeric | Platelet reproducing period is defined as count >50000/mm3. |
| 31. | acute_GvHD_II_III_IV | Boolean | Development of stage II, III, or IV acute graft versus host disease |
| 32. | acute_GvHD_III_IV | Boolean | Stage III or IV growth of acute graft versus host disease |
| 33. | time_to_acute_GvHD_III_IV | Numeric | Time required for the onset of stage III or IV acute graft against host disease |
| 34. | extensive_chronic_GvHD | Categorical | Chronic graft versus host disease develops to a large extent |
| 35. | relapse | Boolean | Disease relapse |
| 36. | survival_time | Numeric | In days, the time of observation or time to event |
| 37. | survival_status | Categorical | Status of survival |

Table 2: Statistical information of numeric attributes.

| Sl. No. | Numeric Attributes | Maximum | Minimum | Mean | Standard Deviation |
|---|---|---|---|---|---|
| 1. | donor_age | 55.5534 | 18.6466 | 33.4721 | 8.2718 |
| 2. | recipient_age | 20.2000 | 0.6000 | 9.9316 | 5.3056 |
| 3. | recipient_body_mass | 103.4000 | 6.0000 | 35.8845 | 19.7522 |
| 4. | CD34_x1e6_per_kg | 57.7800 | 0.7900 | 11.8918 | 9.9144 |
| 5. | CD3_x1e8_per_kg | 20.0200 | 0.0400 | 4.7848 | 3.8530 |
| 6. | CD3_to_CD34_ratio | 99.5610 | 0.2041 | 5.3222 | 9.4806 |
| 7. | ANC_recovery | 1000000.0000 | 9.0000 | 26752.8663 | 161747.2005 |
| 8. | PLT_recovery | 1000000.0000 | 9.0000 | 90937.9198 | 288242.4077 |
| 9. | time_to_acute_GvHD_III_IV | 1000000.0000 | 10.0000 | 775408.0428 | 418425.2527 |
| 10. | survival_time | 3364.0000 | 6.0000 | 938.7433 | 849.5895 |

## 3.2 Chi-squared Test:

As a type of statistical procedure, Chi-squared tests are used to determine the level of independence between categorical variables. It is also a widely used non-parametric method for parametric and normal distribution testing of nominal data [43]. This technique is intended for feature tests that are independent of one another. This produces the Chi-squared score, which is used to identify the most highly correlated feature for ML models to predict desired outcomes [44]. The Chi-squared score indicates the degree to which the attributes of a dataset are related. An attribute with a low score, indicates that it has a very low predictive ability for the dataset's desired outcome column. Therefore, by utilizing this information, the most critical features may be identified, and more efficient models may be deployed on large datasets. The Chi-squared statistical test formula can be written as follows –

$$\chi^2 = \sum \frac{(O-E)^2}{E}, where \begin{cases} O \text{ denotes the observed frequencies} \\ E \text{ denotes the expected frequencies} \end{cases}$$

## 3.3 Hyper Parameter Optimization (HPO):

The parameters that define the architecture of ML models are known as hyper-parameters. Hence, the optimization of hyper-parameters has a substantial impact on the formation of ideal models for certain tasks. While training the model, hyper-parameters are optimized using validation data from a dataset. Typically, Grid Search Cross Validation and Random Search Cross Validation are two HPO processes that work well for a variety of ML tasks [45]. HPO is critical for determining the optimal performance of any ML model because it establishes the model's core architecture [46]. Moreover, the importance of HPO was discovered by several researchers, and is now widely employed in ML-based prediction [47]. The grid search algorithm evaluates all possible combinations from a given set of hyperparameters, whereas the random search algorithm just attempts some random possible combinations [48]. As a result, even though it takes a bit longer than a random search, the grid search technique yields better results when tuning the hyperparameters of any ML algorithm. Hence, the grid search technique is employed in this study to fine-tune the hyperparameters and achieve better results. Figure 2 illustrates the grid search algorithm.

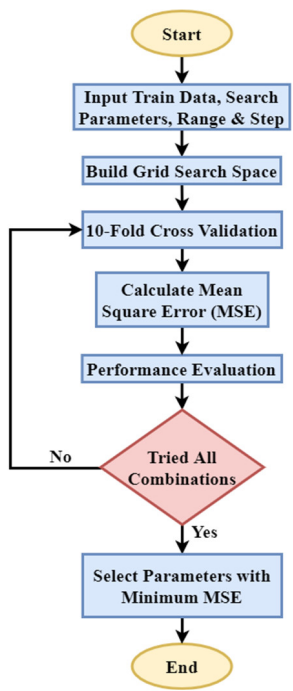

Figure 2: Grid-search algorithm.

### 3.4 Workflow:

Following early data analysis, the dataset underwent multiple preprocessing stages before being used in the machine learning models. First, the dataset underwent multiple preprocessing stages before being used in ML models. The missing values of the dataset were filled with mean values for numerical ones and most frequent values for categorical ones. Since categorical data cannot be handled by ML models, the categorical variables were encoded into numerical form. The dummy variable encoding technique was employed for this purpose, and the attributes were turned into boolean attributes that could readily fit into any ML model [49]. Second, the attributes were then normalized using the standard scaling method to avoid bias from the ML models [50], leaving the dataset with 59 columns after preprocessing. To discover the correlation between attributes, the correlation heatmap is generated using the processed dataset, as depicted in Figure 3. Third, the dataset was split into train and test sets in proportions of 80% and 20%, respectively. Seven ML algorithms: Decision Tree (DT), Random Forest (RF), Logistic Regression (LR), K-Nearest Neighbors (KNN), Gradient Boosting Classifier (GBC), Ada Boost (AdB), XG Boost (XGB), were fed and trained on this dataset, and performance metrics were obtained. Moreover, the Chi-squared statistical test is used to determine the most important features, and the test score is represented in Table 3. Once the chi-squared score is calculated, a minimum number of features are determined that can still predict survival reliably, using fewer electronic health records and computational resources. As a result, the top 11 features were chosen empirically from Table 3 and were analyzed for the prediction of the models. The correlation heatmap using these 11 features is shown in Figure 4.

As mentioned earlier, a total of four distinct experiments, A, B, C, and D, were carried out in this study. In the experiments A and C, no HPO was performed. However, in the experiments B and D, the train dataset was cross validated using Grid Search Cross Validation (GSCV) to determine the optimum hyper-parameters of the ML models. After training the ML models, the test dataset was fed to evaluate the performance of various models. Finally, all performance metrics were calculated, and various comparisons and analyses were performed to determine the impact of hyper-parameter tuning and the use of the full feature dataset and the reduced dataset in which attributes were chosen based on the results of the Chi-squared test. This research was entirely carried out on an *Intel Core i5-8300H* CPU operating at *2.30 GHz*, *8 GB of RAM*, and an *NVIDIA GTX 1050 Ti* graphics unit with *4 GB* of *GPU* memory using *Jupyter Notebook v6.1.4* (*Python 3 v3.8.5*) and *Anaconda-v4.10.3*.

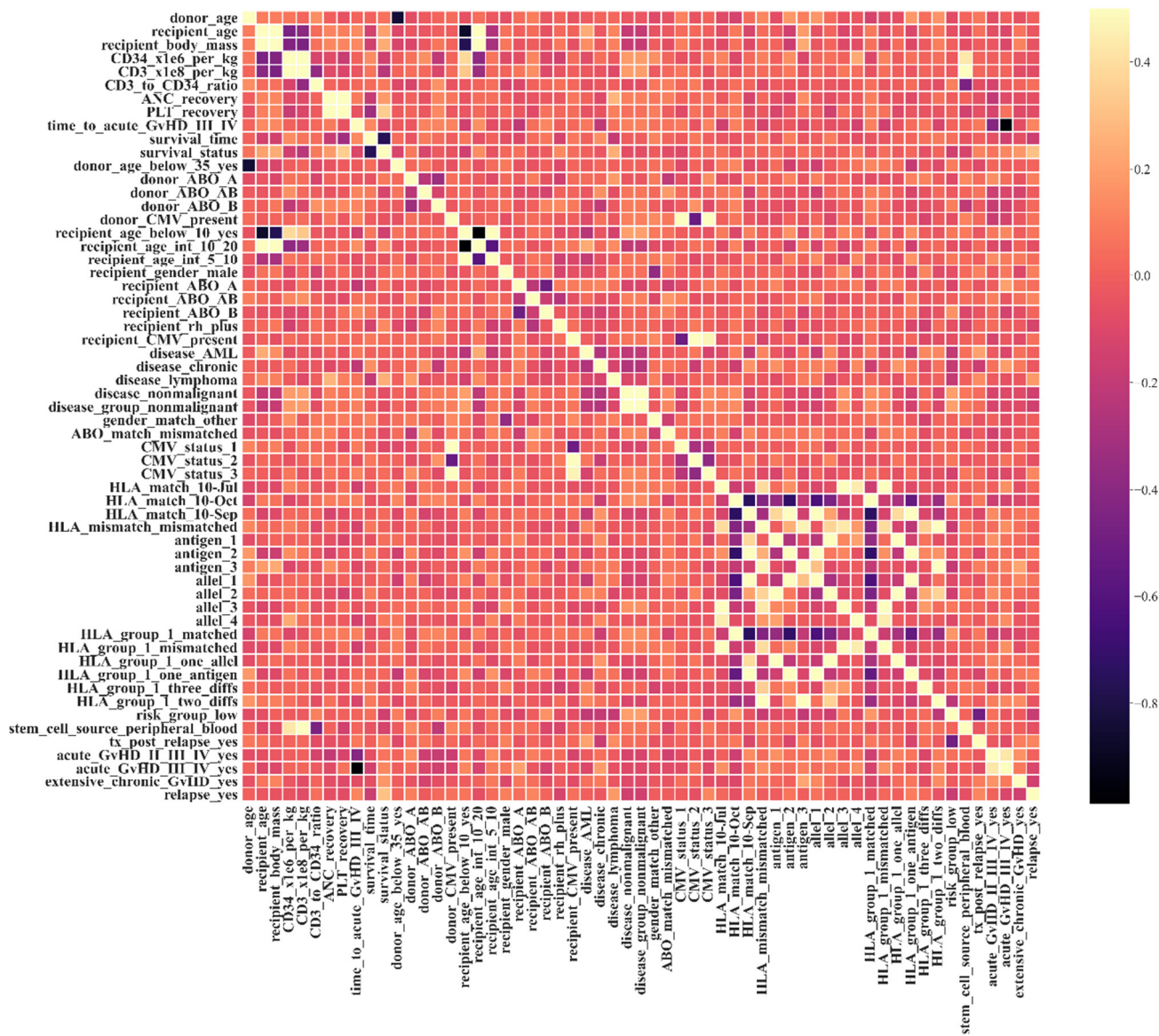

Figure 3: Correlation heatmap (full feature dataset).

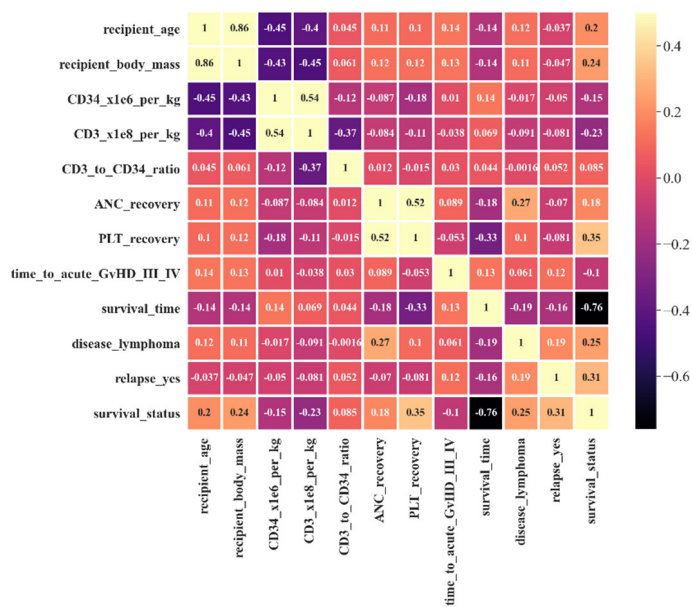

Figure 4: Correlation heatmap (reduced feature dataset).

## 4 RESULTS:

### 4.1 Chi-squared Test Scores:

After preprocessing the data, which includes filling in missing values, encoding categorical variables, and normalization, the Chi-squared statistical test is used to determine the attributes' independence. The summary of the test in this preprocessed dataset is shown in table 3. The top attribute in this list is "PLT recovery", followed by "ANC recovery", "time_to_acute_GvHD_III_IV", "survival_time," and so on.

Table 3: Chi-squared test results.

| Rank | Attribute | Serial in main dataset | Chi-squared Score |
|---|---|---|---|
| 1 | PLT_recovery | 7 | 20390478.3300 |
| 2 | ANC_recovery | 6 | 5996503.1490 |
| 3 | time_to_acute_GvHD_III_IV | 8 | 425033.1475 |
| 4 | survival_time | 9 | 82924.0104 |
| 5 | recipient_body_mass | 2 | 115.2398 |
| 6 | CD34_x1e6_per_kg | 3 | 33.0659 |
| 7 | CD3_x1e8_per_kg | 4 | 30.4617 |
| 8 | CD3_to_CD34_ratio | 5 | 22.9098 |
| 9 | recipient_age | 1 | 21.8122 |
| 10 | relapse_yes | 57 | 15.2012 |
| 11 | disease_lymphoma | 26 | 10.8000 |
| 12 | extensive_chronic_GvHD_yes | 56 | 4.5430 |
| 13 | acute_GvHD_III_IV_yes | 55 | 2.3408 |
| 14 | tx_post_relapse_yes | 53 | 2.2043 |
| 15 | donor_ABO_AB | 12 | 2.1356 |
| 16 | recipient_age_below_10_yes | 15 | 1.9963 |
| 17 | recipient_age_int_10_20 | 16 | 1.9416 |
| 18 | risk_group_low | 51 | 1.5054 |
| 19 | CMV_status_1 | 31 | 1.4080 |

| Rank | Attribute | Serial in main dataset | Chi-squared Score |
|---|---|---|---|
| 20 | donor_age | 0 | 1.3863 |
| 21 | allel_4 | 44 | 1.2000 |
| 22 | donor_age_below_35_yes | 10 | 1.0782 |
| 23 | disease_group_nonmalignant | 28 | 0.8167 |
| 24 | disease_nonmalignant | 27 | 0.8167 |
| 25 | donor_ABO_A | 11 | 0.7892 |
| 26 | HLA_group_1_three_diffs | 49 | 0.6750 |
| 27 | stem_cell_source_peripheral_blood | 52 | 0.6703 |
| 28 | recipient_rh_plus | 22 | 0.4602 |
| 29 | HLA_match_10-Jul | 34 | 0.4267 |
| 30 | HLA_group_1_mismatched | 46 | 0.4267 |
| 31 | antigen_1 | 38 | 0.4063 |
| 32 | antigen_3 | 40 | 0.3857 |
| 33 | recipient_age_int_5_10 | 17 | 0.3765 |
| 34 | acute_GvHD_II_III_IV_yes | 54 | 0.3440 |
| 35 | ABO_match_mismatched | 30 | 0.3380 |
| 36 | recipient_ABO_A | 19 | 0.3197 |
| 37 | donor_CMV_present | 14 | 0.2630 |
| 38 | recipient_ABO_AB | 20 | 0.2564 |
| 39 | CMV_status_3 | 33 | 0.1984 |
| 40 | disease_chronic | 25 | 0.1896 |
| 41 | recipient_CMV_present | 23 | 0.1361 |
| 42 | HLA_match_10-Sep | 36 | 0.1313 |
| 43 | HLA_match_10-Oct | 35 | 0.1280 |
| 44 | HLA_group_1_matched | 45 | 0.1280 |
| 45 | HLA_group_1_one_antigen | 48 | 0.0794 |
| 46 | donor_ABO_B | 13 | 0.0762 |
| 47 | allel_1 | 41 | 0.0733 |
| 48 | antigen_2 | 39 | 0.0611 |
| 49 | allel_3 | 43 | 0.0500 |
| 50 | recipient_gender_male | 18 | 0.0429 |
| 51 | recipient_ABO_B | 21 | 0.0427 |
| 52 | HLA_group_1_one_allel | 47 | 0.0381 |
| 53 | allel_2 | 42 | 0.0375 |
| 54 | CMV_status_2 | 32 | 0.0356 |
| 55 | HLA_group_1_two_diffs | 50 | 0.0281 |
| 56 | HLA_mismatch_mismatched | 37 | 0.0107 |
| 57 | gender_match_other | 29 | 0.0054 |
| 58 | disease_AML | 24 | 0.0000 |

## 4.2 Experiment A: with a full set of features and default hyper-parameters

This experiment was conducted using the processed full-feature dataset with no optimization of hyperparameters. The dataset for this experiment has 58 attributes and 1 objective attribute. Figure 3 shows the correlation heat map for the whole feature dataset. The performance of the ML classifiers was evaluated with default hyper-parameters. The confusion matrices are listed from Table 4 to Table 10, and the Receiver Operating Characteristics (ROC) curve is shown in Figure 5. The ROC curve can be used to discover the ideal ML model for a given task, thus removing suboptimal models. Then table 11 contains accuracy, precision, recall, F1 and ROC_AUC values. The DT, LR, GBC, and AdB have the best accuracy, precision, and F1 score. In this experiment, DT has the highest recall and ROC_AUC.

Table 4: Confusion matrix of Decision Tree.

| **Decision Tree (DT)** | | **Predicted** | |
|---|---|---|---|
| | | **Positive** | **Negative** |
| **Actual** | **True** | 19 | 17 |
| | **False** | 2 | 0 |

Table 5: Confusion matrix of Random Forest.

| **Random Forest (RF)** | | **Predicted** | |
|---|---|---|---|
| | | **Positive** | **Negative** |
| **Actual** | **True** | 19 | 16 |
| | **False** | 2 | 1 |

Table 6: Confusion matrix of Logistic Regression.

| **Logistic Regression (LR)** | | **Predicted** | |
|---|---|---|---|
| | | **Positive** | **Negative** |
| **Actual** | **True** | 20 | 16 |
| | **False** | 1 | 1 |

Table 7: Confusion matrix of K-Nearest Neighbors.

| **K-Nearest Neighbors (KNN)** | | **Predicted** | |
|---|---|---|---|
| | | **Positive** | **Negative** |
| **Actual** | **True** | 18 | 5 |
| | **False** | 3 | 12 |

Table 8: Confusion matrix of Gradient Boosting Classifier.

| **Gradient Boosting Classifier (GBC)** | | **Predicted** | |
|---|---|---|---|
| | | **Positive** | **Negative** |
| **Actual** | **True** | 20 | 16 |
| | **False** | 1 | 1 |

Table 9: Confusion matrix of Ada Boost.

| **Ada Boost (Adb)** | | **Predicted** | |
|---|---|---|---|
| | | **Positive** | **Negative** |
| **Actual** | **True** | 20 | 16 |
| | **False** | 1 | 1 |

Table 10: Confusion matrix of XG Boost.

| **XG Boost (XGB)** | | **Predicted** | |
|---|---|---|---|
| | | **Positive** | **Negative** |
| **Actual** | **True** | 18 | 16 |
| | **False** | 3 | 1 |

Table 11: Performance metrics of ML algorithms by Experiment A.

| Algorithm | Accuracy | Precision | Recall | F1 | ROC_AUC |
|---|---|---|---|---|---|
| Decision Tree | **0.9473** | 0.9047 | **1.000** | 0.9500 | **0.9523** |
| Random Forest | 0.9210 | 0.9047 | 0.9500 | 0.9268 | 0.9229 |
| Logistic Regression | **0.9473** | **0.9523** | 0.9523 | **0.9523** | 0.9467 |
| K-Nearest Neighbors | 0.6052 | 0.8571 | 0.6000 | 0.7058 | 0.5756 |
| Gradient Boosting Classifier | **0.9473** | **0.9523** | 0.9523 | **0.9523** | 0.9467 |
| Ada Boost | **0.9473** | **0.9523** | 0.9523 | **0.9523** | 0.9467 |
| XG Boost | 0.8947 | 0.8571 | 0.9473 | 0.9000 | 0.8991 |

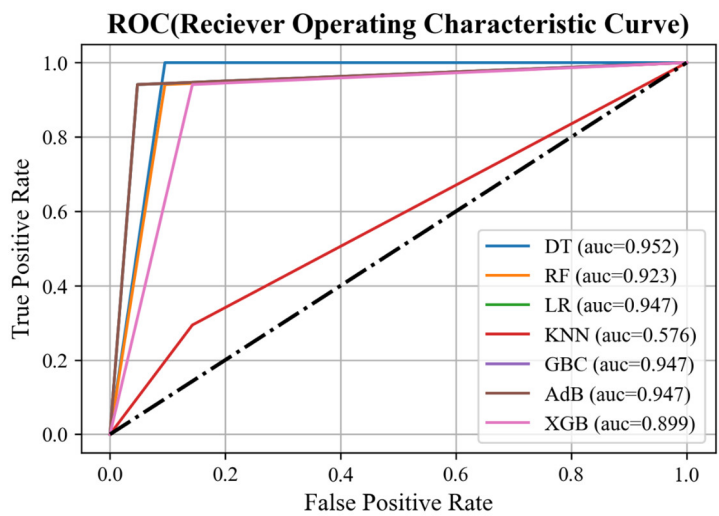

Figure 5: ROC curve for of all the ML models for Experiment A.

### 4.3 Experiment B: with a full set of features and hyper-parameter optimization

In this experiment, the full set of features of the dataset was utilized and the hyper-parameter optimization was also performed using Grid Search Cross-Validation (GSCV). The training dataset was cross-validated 10-folds using GSCV to determine the optimal hyper-parameters and using which all-other performance metrics were assessed. Seven ML algorithms, mentioned earlier, are investigated in this section, and performance metrics are computed. The tables from 12 to 18 provide the confusion matrices for this experiment, and Figure 6 presents the ROC curve. Table 19 depicts all the additional performance metrics, such as accuracy, precision, recall, F1, and ROC_AUC. As seen from this table, the DT, LR, GBC, and AdB algorithms all performed well in this experiment. Moreover, DT outperforms the other algorithms in terms of precision and F1 score, whereas LR, GBC, and AdB have the highest recall and ROC_AUC.

Table 12: Confusion matrix of Decision Tree.

| Decision Tree (DT) | | Predicted | |
|---|---|---|---|
| | | Positive | Negative |
| Actual | True | 21 | 15 |
| | False | 0 | 2 |

Table 13: Confusion matrix of Random Forest.

| Random Forest (RF) | | Predicted | |
|---|---|---|---|
| | | Positive | Negative |
| Actual | True | 19 | 16 |
| | False | 2 | 1 |

Table 14: Confusion matrix of Logistic Regression.

| Logistic Regression (LR) | | Predicted | |
|---|---|---|---|
| | | Positive | Negative |
| Actual | True | 20 | 16 |
| | False | 1 | 1 |

Table 15: Confusion matrix of K-Nearest Neighbors.

| K-Nearest Neighbors (KNN) | | Predicted | |
|---|---|---|---|
| | | Positive | Negative |
| Actual | True | 18 | 8 |
| | False | 3 | 9 |

Table 16: Confusion matrix of Gradient Boosting Classifier.

| Gradient Boosting Classifier (GBC) | | Predicted | |
|---|---|---|---|
| | | Positive | Negative |
| Actual | True | 20 | 16 |
| | False | 1 | 1 |

Table 17: Confusion matrix of Ada Boost.

| Ada Boost (Adb) | | Predicted | |
|---|---|---|---|
| | | Positive | Negative |
| Actual | True | 20 | 16 |
| | False | 1 | 1 |

Table 18: Confusion matrix of XG Boost.

| XG Boost (XGB) | | Predicted | |
|---|---|---|---|
| | | Positive | Negative |
| Actual | True | 19 | 16 |
| | False | 2 | 1 |

Table 19: Performance metrics of ML algorithms by Experiment B.

| Algorithm | Accuracy | Precision | Recall | F1 | ROC_AUC |
|---|---|---|---|---|---|
| Decision Tree | **0.9473** | **1.0000** | 0.9130 | **0.9545** | 0.9411 |
| Random Forest | 0.9210 | 0.9047 | 0.9500 | 0.9268 | 0.9229 |
| Logistic Regression | **0.9473** | 0.9523 | **0.9523** | 0.9523 | **0.9467** |
| K-Nearest Neighbors | 0.6842 | 0.8571 | 0.6666 | 0.7500 | 0.6638 |
| Gradient Boosting Classifier | **0.9473** | 0.9523 | **0.9523** | 0.9523 | **0.9467** |
| Ada Boost | **0.9473** | 0.9523 | **0.9523** | 0.9523 | **0.9467** |
| XG Boost | 0.9210 | 0.9047 | 0.9500 | 0.9268 | 0.9229 |

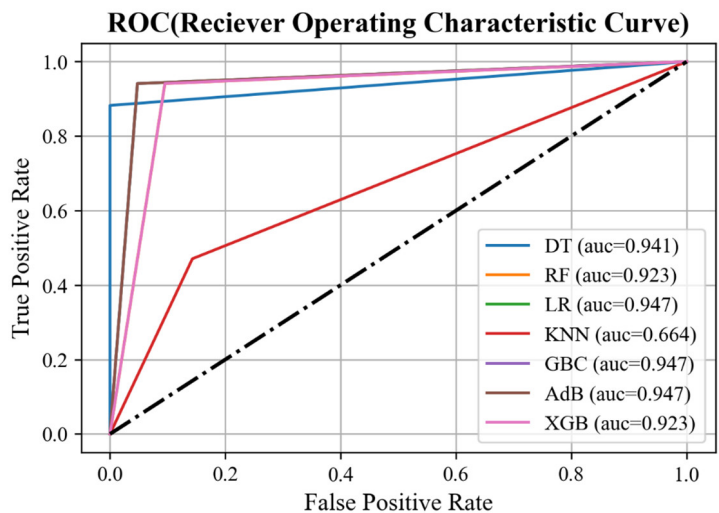

Figure 6: ROC curve for of all the ML models for Experiment B.

## 4.4 Experiment C: A reduced dataset based on Chi-squared test and default hyper-parameters

This experiment was conducted based on the results of the Chi-squared test. Following data preprocessing, Chi-squared feature ranking is performed, and the resulting scores are summarized in Table 3. According to that table, the first eleven features are considered in this experiment since they have a higher Chi-squared score. Hence, ML models were initially trained on the selected training set and subsequently verified on the test set without any HPO. For this experiment, the confusion matrices are shown from Table 20 to Table 26. However, the ROC curve is presented in Figure 7. The performance metrics are reported in Table 27, and it is apparent that the KNN surpasses the rest of the classifiers regarding accuracy, F1score, recall, and the ROC_AUC value. However, when it comes to precision, RF, LR, KNN, GNB, and XGB all do well.

Table 20: Confusion matrix of Decision Tree.

| Decision Tree (DT) | | Predicted | |
|---|---|---|---|
| | | Positive | Negative |
| Actual | True | 17 | 14 |
| | False | 4 | 3 |

Table 21: Confusion matrix of Random Forest.

| Random Forest (RF) | | Predicted | |
|---|---|---|---|
| | | Positive | Negative |
| Actual | True | 19 | 12 |
| | False | 2 | 5 |

Table 22: Confusion matrix of Logistic Regression.

| Logistic Regression (LR) | | Predicted | |
|---|---|---|---|
| | | Positive | Negative |
| Actual | True | 19 | 14 |
| | False | 2 | 3 |

Table 23: Confusion matrix of K-Nearest Neighbors.

| K-Nearest Neighbors (KNN) | | Predicted | |
|---|---|---|---|
| | | Positive | Negative |
| Actual | True | 19 | 16 |
| | False | 2 | 1 |

Table 24: Confusion matrix of Gradient Boosting Classifier.

| Gradient Boosting Classifier (GBC) | | Predicted | |
|---|---|---|---|
| | | Positive | Negative |
| Actual | True | 19 | 12 |
| | False | 2 | 5 |

Table 25: Confusion matrix of Ada Boost.

| Ada Boost (AdB) | | Predicted | |
|---|---|---|---|
| | | Positive | Negative |
| Actual | True | 18 | 12 |
| | False | 3 | 5 |

Table 26: Confusion matrix of XG Boost.

| XG Boost (XGB) | | Predicted | |
|---|---|---|---|
| | | Positive | Negative |
| Actual | True | 19 | 12 |
| | False | 2 | 5 |

Table 27: Performance metrics of ML algorithms by Experiment C.

| Algorithm | Accuracy | Precision | Recall | F1 | ROC_AUC |
|---|---|---|---|---|---|
| Decision Tree | 0.8157 | 0.8095 | 0.8500 | 0.8292 | 0.8165 |
| Random Forest | 0.8157 | **0.9047** | 0.7916 | 0.8444 | 0.8053 |
| Logistic Regression | 0.8684 | **0.9047** | 0.8636 | 0.8837 | 0.8641 |
| K-Nearest Neighbors | **0.9210** | **0.9047** | **0.9500** | **0.9268** | **0.9229** |
| Gradient Boosting Classifier | 0.8157 | **0.9047** | 0.7916 | 0.8444 | 0.8053 |
| Ada Boost | 0.7894 | 0.8571 | 0.7826 | 0.8181 | 0.7815 |
| XG Boost | 0.8157 | **0.9047** | 0.7916 | 0.8444 | 0.8053 |

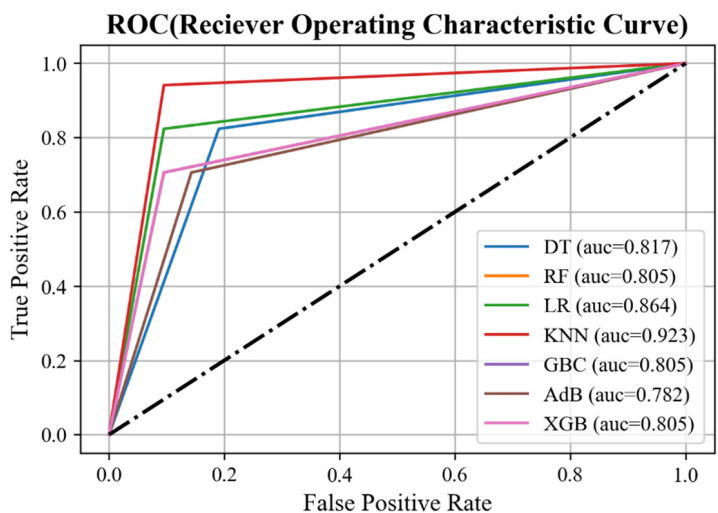

Figure 7: ROC curve for of all the ML models for Experiment C.

## 4.5 Experiment D: A reduced dataset based on Chi-squared test and hyper-parameter optimization)

Like the previous one, experiment D makes use of a reduced feature dataset based on the Chi-squared test, and HPO is performed using GSCV with 10-fold cross validation, and the optimal hyper-parameters are determined and used to evaluate performance metrics. The reduced dataset comprises 11 attributes, and the performances are evaluated based on them.

As before, multiple ML methods such as DT, RF, LR, KNN, GBC, AdB, and XGB are deployed. From Table 28 to Table 34, the confusion matrices are shown, and Figure 8 illustrates the ROC curve for these tests. Table 35 illustrates that the DT outperforms all other algorithms in every performance metric. However, in terms of precision, the DT and KNN perform the best altogether.

Table 28: Confusion matrix of Decision Tree.

| **Decision Tree (DT)** | | **Predicted** | |
|---|---|---|---|
| | | Positive | Negative |
| **Actual** | True | 20 | 16 |
| | False | 1 | 1 |

Table 29: Confusion matrix of Random Forest.

| **Random Forest (RF)** | | **Predicted** | |
|---|---|---|---|
| | | Positive | Negative |
| **Actual** | True | 19 | 13 |
| | False | 2 | 4 |

Table 30: Confusion matrix of Logistic Regression.

| **Logistic Regression (LR)** | | **Predicted** | |
|---|---|---|---|
| | | Positive | Negative |
| **Actual** | True | 19 | 14 |
| | False | 2 | 3 |

Table 31: Confusion matrix of K-Nearest Neighbors.

| **K-Nearest Neighbors (KNN)** | | **Predicted** | |
|---|---|---|---|
| | | Positive | Negative |
| **Actual** | True | 20 | 14 |
| | False | 1 | 3 |

Table 32: Confusion matrix of Gradient Boosting Classifier.

| **Gradient Boosting Classifier (GBC)** | | **Predicted** | |
|---|---|---|---|
| | | Positive | Negative |
| **Actual** | True | 19 | 12 |
| | False | 2 | 5 |

Table 33: Confusion matrix of Ada Boost.

| **Ada Boost (Adb)** | | **Predicted** | |
|---|---|---|---|
| | | Positive | Negative |
| **Actual** | True | 19 | 12 |
| | False | 2 | 5 |

Table 34: Confusion matrix of XG Boost.

| **XG Boost (XGB)** | | **Predicted** | |
|---|---|---|---|
| | | Positive | Negative |
| **Actual** | True | 19 | 12 |
| | False | 2 | 5 |

Table 35: Performance metrics of ML algorithms by Experiment D.

| Algorithm | Accuracy | Precision | Recall | F1 | ROC_AUC |
|---|---|---|---|---|---|
| Decision Tree | **0.9473** | **0.9523** | **0.9523** | **0.9523** | **0.9467** |
| Random Forest | 0.8421 | 0.9047 | 0.8260 | 0.8636 | 0.8347 |
| Logistic Regression | 0.8684 | 0.9047 | 0.8636 | 0.8837 | 0.8641 |
| K-Nearest Neighbors | 0.8947 | **0.9523** | 0.8695 | 0.9090 | 0.8879 |
| Gradient Boosting Classifier | 0.8157 | 0.9047 | 0.7916 | 0.8444 | 0.8053 |
| Ada Boost | 0.8157 | 0.9047 | 0.7916 | 0.8444 | 0.8053 |
| XG Boost | 0.8157 | 0.9047 | 0.7916 | 0.8444 | 0.8053 |

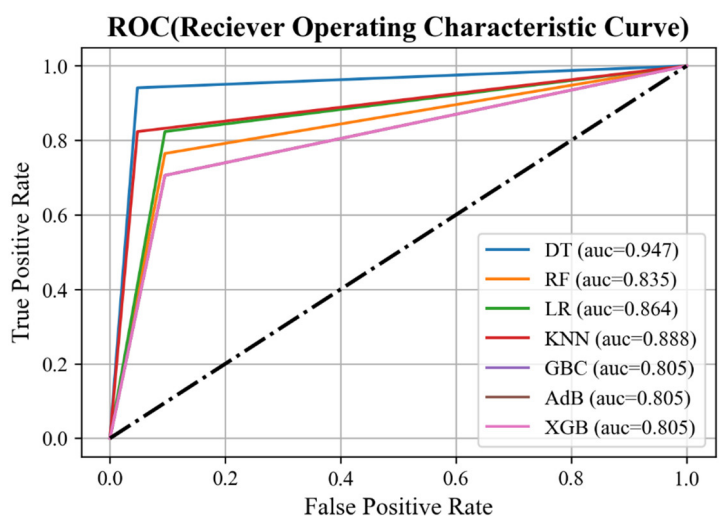

Figure 8 ROC curve for of all the ML models for Experiment D.

## 5 DISCUSSION:

As can be seen, the overall study included four experiments with four different approaches. The whole feature dataset was employed in experiment A without HPO, and the maximum accuracy was found to be 94.73%, as were the precision (0.9523), recall (1), F1 score (0.9523), and ROC_AUC (0.953). In terms of accuracy, the best algorithms are DT, LR, GBC, and AdB. However, in experiment B, the maximum accuracy was 94.73%, and the precision, recall, F1 score, and ROC_AUC were 1, 0.9523, 0.9545, and 0.9467, respectively. The well-performing algorithms were the same as in experiment A, but the overall performance was improved in this experiment since HPO was performed with 10-folds GSCV method. On the other hand, the experiments C and D are carried out based on the Chi-squared test results. The top 11 features from the dataset were extracted, and this reduced dataset was employed in these tests. In experiment C, the maximum accuracy of 92.1% was obtained for KNN, with precision (0.9047), recall (0.95), F1 score (0.9268), and ROC_AUC (0.9229). In experiment D, HPO was performed on the same reduced dataset as in experiment C. This time, the performance of all the ML algorithms improved significantly, DT having the best performance measures. The performance measures for DT in experiment D are as follows: accuracy (0.9473), precision (0.9523), recall (0.9523), F1 score (0.9523), and ROC_AUC (0.9467).

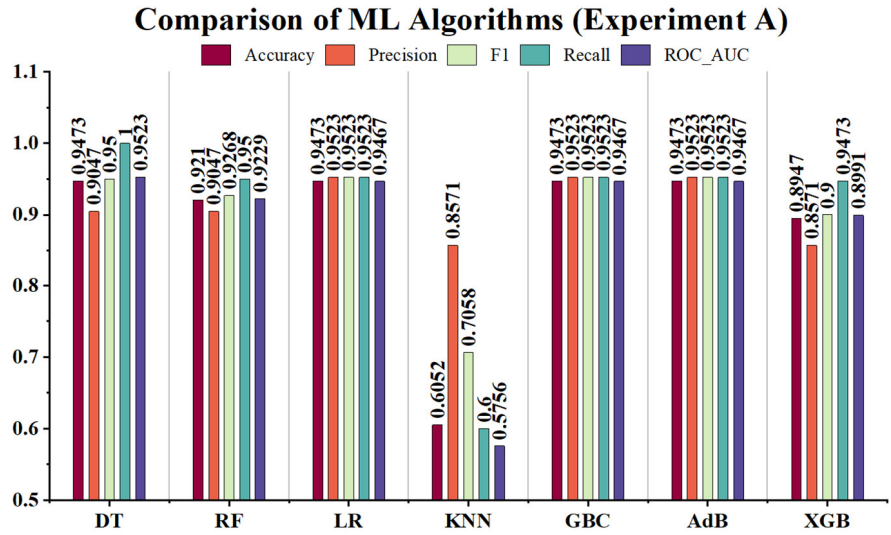

Figure 9: Comparative analysis of the performance metrics for Experiment A.

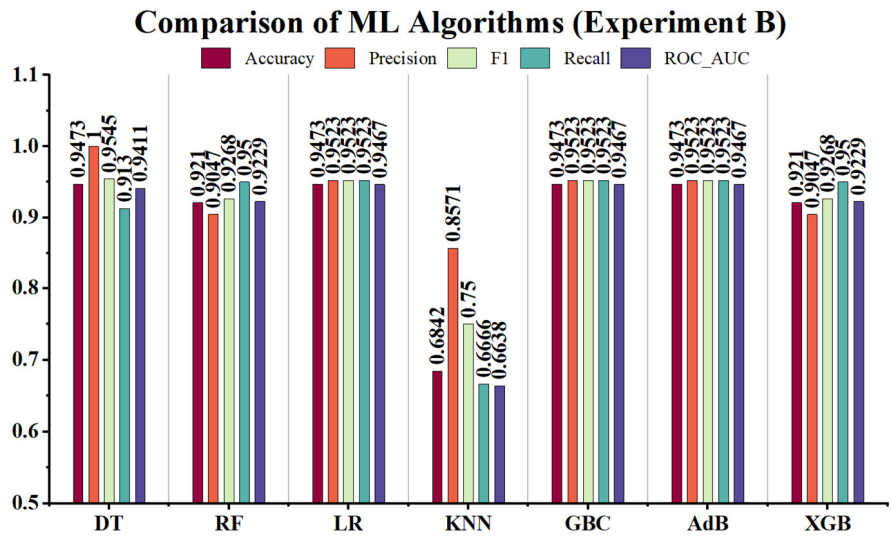

Figure 10: Comparative analysis of the performance metrics for Experiment B.

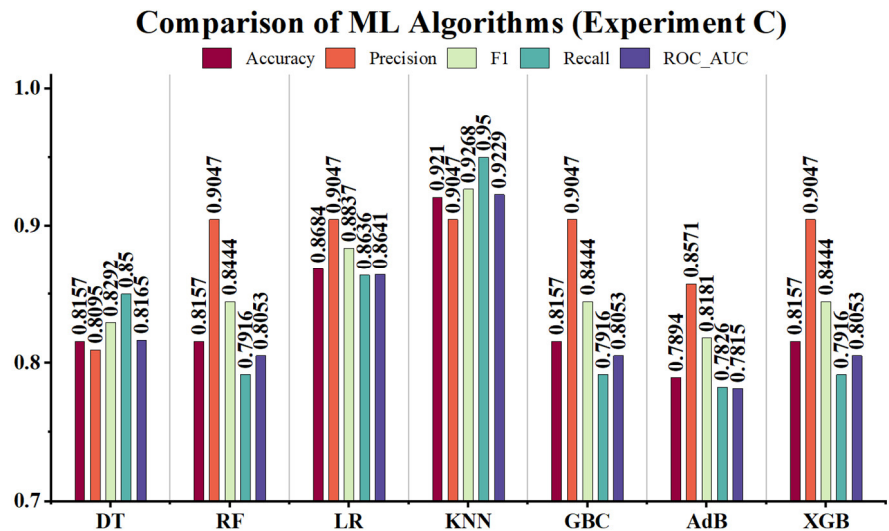

Figure 11: Comparative analysis of the performance metrics for Experiment C.

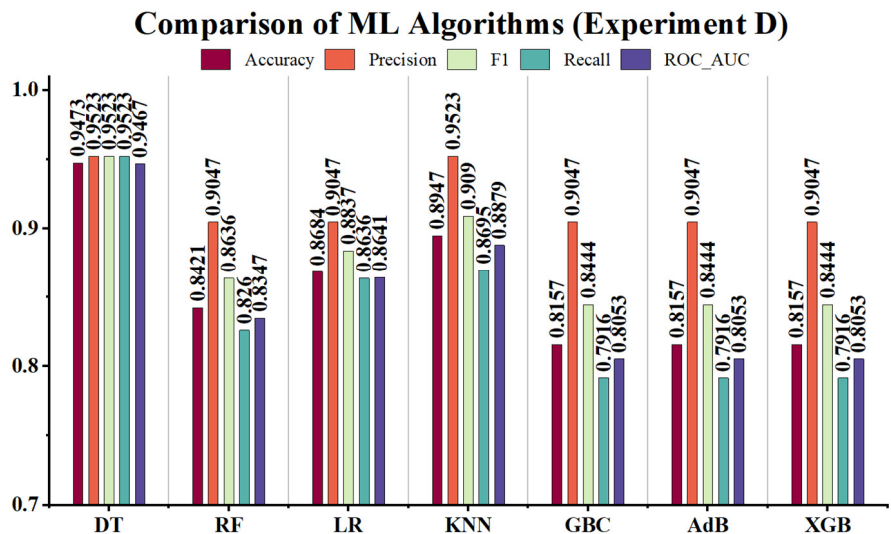

Figure 12 Comparative analysis of the performance metrics for Experiment D.

Based on the above four experiments, it is evident that HPO is critical to enhancing the performance of the ML algorithms, and the Chi-squared test plays a significant role in determining the most important feature. The computation time of GSCV using complete and reduced feature datasets is shown in Table 36 and is displayed using a bar plot in Figure 13. Most of the ML classifiers required less time in the reduced feature dataset without significantly affecting performance, which is an encouraging result of our study. The comparison between the four experiments is shown in Figure 14 in terms of accuracy, precision, recall, F1, and ROC_AUC.

The top five critical attributes established in this study are: "PLT recovery", "ANC recovery", "duration to acute GvHD III IV", "survival time", and "recipient body mass". As can be seen, we found maximum accuracy (0.9473), precision (1), recall (1), F1 score (0.9545), and AUC (0.9523).

Table 36: Comparison of computational time.

| Algorithm | Computation Time for Full Feature Data (seconds) | Computation Time for Reduced Feature Data (seconds) |
|---|---|---|
| Decision Tree | 17.17 | 19.26 |
| Random Forest | 118.00 | 161.35 |
| Logistic Regression | 35.56 | **26.74** |
| K-Nearest Neighbors | 7.17 | **5.92** |
| Gradient Boosting Classifier | 52.19 | **39.00** |
| Ada Boost | 14.46 | **17.00** |
| XG Boost | 586.00 | **550.00** |

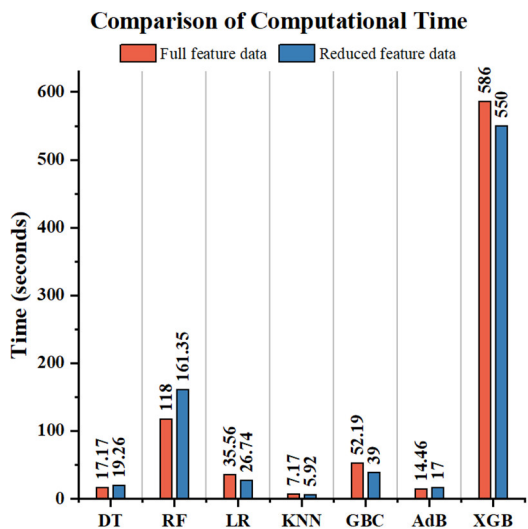

Figure 13: Comparison of computational time (between Experiment B and D).

Table 37: Performance metrics of four experiments (maximum).

| Experiment | Accuracy | Precision | Recall | F1 Score | ROC_AUC |
|---|---|---|---|---|---|
| A | 0.9473 | 0.9523 | 1.0000 | 0.9523 | 0.9523 |
| B | 0.9473 | 1.0000 | 0.9523 | 0.9545 | 0.9467 |
| C | 0.9210 | 0.9047 | 0.9500 | 0.9268 | 0.9229 |
| D | 0.9473 | 0.9523 | 0.9523 | 0.9523 | 0.9467 |

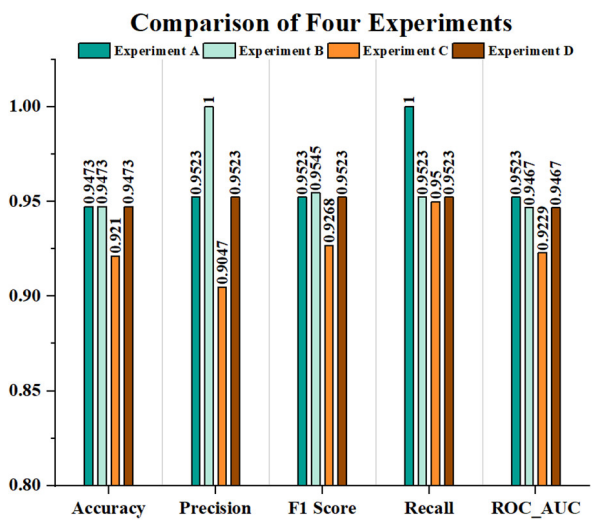

Figure 14: Comparison of performance metrics (Experiment A, B, C, and D).

Previously, researchers employed a prediction method to predict the survivability of patients receiving BMTs. For instance, Adam Gudys et al. employed a rule-based predictive model in this dataset and produced a tool named RuleKit for predicting BMT survival rates in children [51]. Likewise, M. Sikora et al. established a framework that is based on decision rules and the rule induction approach [52]. In a similar study, K. Karami et al. combined ML and feature selection methods to identify the most appropriate factors for predicting AML patient survival [53]. They used six ML algorithms, like DT, RF, LR, Naive Bayes, W-Bayes Net, and Gradient Boosted Tree (GBT). With an AUC value of 0.930, the GBT was found to have an 86.17% accuracy, making it the most accurate predictor of AML patient survival using the Relief algorithm for feature selection. Moreover, V. Leclerc et al. employed a Tree-Augmented Nave Bayesian network to develop a certified decision support tool for selecting the most suitable initial dose of intravenous Cyclosporine A (CsA) in pediatric patients undergoing

HSCT [54]. A ten-year monocentric dataset was used after discretization using Shannon entropy and equal width intervals. The AUC-ROC of the TAN Bayesian model is 0.804 on average, with a 32.8% misclassification rate and true-positive and false-positive rates of 0.672 and 0.285, respectively. Additionally, R. Bortnick et al. investigated the outcomes of 65 patients with Myelodysplastic Syndromes (MDS) in infancy who had received HSCT and had a germline GATA2 mutation (GATA2mut) [55]. Overall survival was found to be 75% after five years, while Disease-Free Survival (DFS) was 70%. On the other hand, V. Hazar et al. evaluated the results of 62 pediatric patients who received HSCT for relapsed Non-Hodgkin Lymphoma (rr-NHL). The Overall Survival (OS) rate was determined to be 65%, whereas the Event-Free Survival (EFS) rate was found to be 48% [56]. However, J. Qi et al. used cox proportional hazard to assess bleeding's independent prognostic value and fine-gray competing risk models for survival analyses; lasso regression to select a training set to derive the bleeding score; and logistic regression to derive the value-added score. There was an increased cumulative incidence of overall mortality (HR = 10.90), non-relapse mortality (HR = 14.84), and combined endpoints (HR = 9.30), but not the cumulative incidence of relapse in higher bleeding class HSCT patients [57]. S. Averill et al. examined the Dinakara and Cotes equations' capacity to predict post-HSCT pulmonary problems and death using paired t-tests, ANOVA models, and LR on a dataset derived from chart reviews of patients having HSCT for the first time at Riley Hospital for Children, employing a dataset kept by the Pediatric Stem Cell Transplant Program. The ability of the Dinakara and Cotes equations to predict death and pulmonary issues after HSCT was similar [58]. Further, in 130 patients receiving melphalan 200 mg/m$^2$ and Autologous Stem Cell Transplant (ASCT), K. Sweiss et al. examined the impact of clinically obtainable immune-related indicators on Treatment-Free Survival (TFS) [59]. Moreover, Y. M. Liu et al. investigated the relationship between Health-Related Quality of Life (HRQoL) and duration of hospital stay in pediatric patients receiving allogeneic HSCT, as well as 1-year survival [60]. Furthermore, on behalf of the European Society for Blood and Marrow Transplantation, S. Corbacioglu et al. recommended improved diagnostic and severity criteria for SOS/VOD in children [61]. L. Stern et al. reviewed the present state of knowledge on immunological reconstitution in HSCT recipients and discussed the latest mass cytometry research that has made significant contributions to the field [62].

Table 38: Performance comparison of our methodology with state-of-the-art ones.

| References | Authors | Method | Findings |
|---|---|---|---|
| [53] | K. Karami et al. | ML with feature selection | accuracy of 85.17%, and AUC of 0.930 |
| [54] | V. Leclerc et al. | Tree-Augmented Naïve Bayesian network | AUC-ROC of 0.804, 32.8% of misclassified patients |
| [56] | V. Hazar et al. | Kaplan–Meier method & $\chi 2$ test | overall survival (OS) of 65% and event-free survival (EFS) rate of 48% |
| [57] | Y. T. Jiaqian Qi et al. | Cox proportional hazard model & fine-gray competing risk model | Overall mortality (HR = 10.90), non-relapse mortality (HR = 14.84), and combined endpoints (HR = 9.30) |
| This work | | ML with Chi-Squared test | Survival prediction accuracy of 94.73% |

## 6 CONCLUSION AND FUTURE WORKS:

A Bone Marrow Transplant is a crucial life-saving treatment for a certain type of malignancy. For this reason, early detection of survivability after BMT can play a vital role in the patient's treatment process. Moreover, if healthcare providers have a prior prediction, they can make more informed decisions about treatment options. In this regard, technologies like ML can be useful, since they can be used in situations requiring prediction and can uncover hidden patterns in previous data in order to create an accurate prediction. Nowadays, it is increasingly being employed in every situation that requires prediction. In this study, we developed a Chi-squared feature selection method and an HPO based efficient model for predicting the survival of children who received BMT and identified the most significant parameter for survival after BMT. All four experiments that were conducted yielded satisfactory predictions. The models operate well on a synthetic dataset that has been constructed from the raw dataset via a series of preprocessing phases that reduce the dataset's dimensionality. With the entire feature synthetic dataset, experiment A achieves an accuracy of 94.73%. However, as experiment B optimizes the hyper-parameters using the same dataset as experiment A, it achieves the highest overall performance of all the models. On the other hand, experiments C and D make use of the 11 most correlated feature dataset based on the Chi-squared test, and experiment D outperforms all performance measures when combined with HPO, achieving high accuracy (94.73%) with less time, data, and resource consumption. In this study, we obtained maximum accuracy (0.9473), precision (1), recall (1),

F1 (0.9545), AUC (0.9523), and the top five attributes that influence the survivability rate are "PLT recovery," "ANC recovery," "duration to acute GvHD III IV," "survival time," and "recipient body mass." Historically, this dataset has not been evaluated in such a manner before, and it could provide the health sector with a unique perspective. Therefore, this study can make a noteworthy contribution to the development of ML-based healthcare prediction systems in environments where resources are scarce and healthcare practitioners lack more data. Additional research on this topic can be conducted by incorporating deep learning and neural networks. From Bangladesh's perspective, new data can also be gathered from medical records. Moreover, programs or user interfaces can be developed to make this more useful to healthcare professionals.

**DECLARATION OF INTEREST**

**FUNDING INFORMATION**

The authors did not receive any funding for this study.

**References:**

[1] R. L. Siegel, K. D. Miller, and A. Jemal, "Cancer statistics, 2020," *CA. Cancer J. Clin.*, vol. 70, no. 1, pp. 7–30, Jan. 2020, doi: 10.3322/CAAC.21590.

[2] "Understanding Cancer: Metastasis, Stages of Cancer, and More." https://www.medicinenet.com/cancer_101_pictures_slideshow/article.htm (accessed Aug. 31, 2021).

[3] "What is a Bone Marrow Transplant (Stem Cell Transplant)? | Cancer.Net." https://www.cancer.net/navigating-cancer-care/how-cancer-treated/bone-marrowstem-cell-transplantation/what-bone-marrow-transplant-stem-cell-transplant (accessed Aug. 31, 2021).

[4] "Bone Marrow Transplant: Types, Procedure & Risks." https://www.healthline.com/health/bone-marrow-transplant#preparation (accessed Aug. 31, 2021).

[5] Y. M and G. T, "Early decisions in lymphoid development," *Curr. Opin. Immunol.*, vol. 19, no. 2, pp. 123–128, Apr. 2007, doi: 10.1016/J.COI.2007.02.007.

[6] R. T, "Regulation of hematopoietic stem cell self-renewal," *Recent Prog. Horm. Res.*, vol. 58, pp. 283–295, 2003, doi: 10.1210/RP.58.1.283.

[7] M. T. De la Morena and R. A. Gatti, "A History of Bone Marrow Transplantation," *Hematol. Oncol. Clin. North Am.*, vol. 25, no. 1, pp. 1–15, 2011, doi: 10.1016/j.hoc.2010.11.001.

[8] "Big gains in bone marrow transplant survival since mid-2000s." https://www.fredhutch.org/en/news/center-news/2020/01/survival-gains-bone-marrow-transplant.html (accessed Aug. 31, 2021).

[9] E. Simpson and F. Dazzi, "Bone marrow transplantation 1957-2019," *Front. Immunol.*, vol. 10, no. JUN, pp. 1–6, 2019, doi: 10.3389/fimmu.2019.01246.

[10] "What is a Bone Marrow Transplant? | Be The Match." https://bethematch.org/patients-and-families/about-transplant/what-is-a-bone-marrow-transplant-/ (accessed Aug. 31, 2021).

[11] "Bone Marrow Transplantation | Johns Hopkins Medicine." https://www.hopkinsmedicine.org/health/treatment-tests-and-therapies/bone-marrow-transplantation (accessed Aug. 31, 2021).

[12] "Bone Marrow Transplant: Preparation, Procedure, Risks, and Recovery." https://www.webmd.com/cancer/multiple-myeloma/bone-marrow-transplants (accessed Aug. 31, 2021).

[13] "Bone marrow transplant: MedlinePlus Medical Encyclopedia." https://medlineplus.gov/ency/article/003009.htm (accessed Aug. 31, 2021).

[14] "Donation and Transplantation Statistics | Blood Stem Cell." https://bloodstemcell.hrsa.gov/data/donation-and-transplantation-statistics (accessed Aug. 31, 2021).

[15] "Blood and Marrow Transplant Statistics and Outcomes." https://www.chp.edu/our-services/blood-marrow-transplant-cellular-therapies/statistics-outcomes (accessed Aug. 31, 2021).

[16] S. Uddin, A. Khan, M. E. Hossain, and M. A. Moni, "Comparing different supervised machine learning algorithms for disease prediction," *BMC Med. Inform. Decis. Mak.*, vol. 19, no. 1, pp. 1–16, 2019, doi: 10.1186/s12911-019-1004-8.

[17]   R. Bharti, A. Khamparia, M. Shabaz, G. Dhiman, S. Pande, and P. Singh, "Prediction of Heart Disease Using a Combination of Machine Learning and Deep Learning," *Comput. Intell. Neurosci.*, vol. 2021, 2021, doi: 10.1155/2021/8387680.

[18]   P. Ghosh *et al.*, "Efficient prediction of cardiovascular disease using machine learning algorithms with relief and lasso feature selection techniques," *IEEE Access*, vol. 9, pp. 19304–19326, 2021, doi: 10.1109/ACCESS.2021.3053759.

[19]   M. Jiang, Y. Li, C. Jiang, L. Zhao, X. Zhang, and P. E. Lipsky, "Machine Learning in Rheumatic Diseases," *Clin. Rev. Allergy Immunol.*, vol. 60, no. 1, pp. 96–110, 2021, doi: 10.1007/s12016-020-08805-6.

[20]   B. Kurian and V. L. Jyothi, "Breast cancer prediction using an optimal machine learning technique for next generation sequences," *Concurr. Eng. Res. Appl.*, vol. 29, no. 1, pp. 49–57, 2021, doi: 10.1177/1063293X21991808.

[21]   A. S. Kwekha-Rashid, H. N. Abduljabbar, and B. Alhayani, "Coronavirus disease (COVID-19) cases analysis using machine-learning applications," *Appl. Nanosci.*, no. 0123456789, 2021, doi: 10.1007/s13204-021-01868-7.

[22]   S. Vatansever *et al.*, "Artificial intelligence and machine learning-aided drug discovery in central nervous system diseases: State-of-the-arts and future directions," *Med. Res. Rev.*, vol. 41, no. 3, pp. 1427–1473, 2021, doi: 10.1002/med.21764.

[23]   A. Moncada-Torres, M. C. van Maaren, M. P. Hendriks, S. Siesling, and G. Geleijnse, "Explainable machine learning can outperform Cox regression predictions and provide insights in breast cancer survival," *Sci. Rep.*, vol. 11, no. 1, pp. 1–13, 2021, doi: 10.1038/s41598-021-86327-7.

[24]   M. C. F. Cysouw *et al.*, "Machine learning-based analysis of [18F]DCFPyL PET radiomics for risk stratification in primary prostate cancer," *Eur. J. Nucl. Med. Mol. Imaging*, vol. 48, no. 2, pp. 340–349, 2021, doi: 10.1007/s00259-020-04971-z.

[25]   S. Shanthi and N. Rajkumar, "Lung Cancer Prediction Using Stochastic Diffusion Search (SDS) Based Feature Selection and Machine Learning Methods," *Neural Process. Lett.*, vol. 53, no. 4, pp. 2617–2630, 2021, doi: 10.1007/s11063-020-10192-0.

[26]   "UCI Machine Learning Repository: Bone marrow transplant: children Data Set." https://archive.ics.uci.edu/ml/datasets/Bone+marrow+transplant%3A+children (accessed Aug. 25, 2021).

[27]   X. Jin, A. Xu, R. Bie, and P. Guo, "Machine learning techniques and chi-square feature selection for cancer classification using SAGE gene expression profiles," *Lect. Notes Comput. Sci. (including Subser. Lect. Notes Artif. Intell. Lect. Notes Bioinformatics)*, vol. 3916 LNBI, pp. 106–115, 2006, doi: 10.1007/11691730_11.

[28]   N. Radakovich, M. Nagy, and A. Nazha, "Machine learning in haematological malignancies," *Lancet Haematol.*, vol. 7, no. 7, pp. e541–e550, 2020, doi: 10.1016/S2352-3026(20)30121-6.

[29]   V. Gupta, T. M. Braun, M. Chowdhury, M. Tewari, and S. W. Choi, "A Systematic Review of Machine Learning Techniques in Hematopoietic Stem Cell Transplantation (HSCT)," *Sensors 2020, Vol. 20, Page 6100*, vol. 20, no. 21, p. 6100, Oct. 2020, doi: 10.3390/S20216100.

[30]   L. Buturovic *et al.*, "Evaluation of a Machine Learning-Based Prognostic Model for Unrelated Hematopoietic Cell Transplantation Donor Selection," *Biol. Blood Marrow Transplant.*, vol. 24, no. 6, pp. 1299–1306, 2018, doi: 10.1016/j.bbmt.2018.01.038.

[31]   B. R. Logan *et al.*, "Optimal Donor Selection for Hematopoietic Cell Transplantation Using Bayesian Machine Learning," *JCO Clin. Cancer Informatics*, no. 5, pp. 494–507, 2021, doi: 10.1200/cci.20.00185.

[32]   A. Sivasankaran, E. Williams, M. Albrecht, G. E. Switzer, V. Cherkassky, and M. Maiers, "Machine Learning Approach to Predicting Stem Cell Donor Availability," *Biol. Blood Marrow Transplant.*, vol. 24, no. 12, pp. 2425–2432, 2018, doi: 10.1016/j.bbmt.2018.07.035.

[33]   Y. Li *et al.*, "Predicting the Availability of Hematopoietic Stem Cell Donors Using Machine Learning," *Biol. Blood Marrow Transplant.*, vol. 26, no. 8, pp. 1406–1413, 2020, doi: 10.1016/j.bbmt.2020.03.026.

[34]   R. Shouval, O. Bondi, H. Mishan, A. Shimoni, R. Unger, and A. Nagler, "Application of machine learning algorithms for clinical predictive modeling: A data-mining approach in SCT," *Bone Marrow Transplant.*, vol. 49, no. 3, pp. 332–337, 2014, doi: 10.1038/bmt.2013.146.

[35]   J. N. Eckardt, M. Bornhäuser, K. Wendt, and J. M. Middeke, "Application of machine learning in the management of acute myeloid leukemia: Current practice and future prospects," *Blood Adv.*, vol. 4, no. 23, pp. 6077–6085, 2020, doi: 10.1182/bloodadvances.2020002997.

[36]   L. Pan *et al.*, "Machine learning applications for prediction of relapse in childhood acute lymphoblastic leukemia," *Sci. Rep.*, vol. 7, no. 1, pp. 1–9, 2017, doi: 10.1038/s41598-017-07408-0.

[37]   K. Fuse *et al.*, "Patient-based prediction algorithm of relapse after allo-HSCT for acute Leukemia and its usefulness in the decision-making process using a machine learning approach," *Cancer Med.*, vol. 8, no. 11, pp. 5058–5067, 2019, doi: 10.1002/cam4.2401.

[38]  Y. Arai *et al.*, "Using a machine learning algorithm to predict acute graft-versus-host disease following allogeneic transplantation," *Blood Adv.*, vol. 3, no. 22, pp. 3626–3634, 2019, doi: 10.1182/bloodadvances.2019000934.

[39]  D. Cilloni *et al.*, "Transplantation Induces Profound Changes in the Transcriptional Asset of Hematopoietic Stem Cells: Identification of Specific Signatures Using Machine Learning Techniques," *J. Clin. Med.*, vol. 9, no. 6, p. 1670, 2020, doi: 10.3390/jcm9061670.

[40]  R. Shouval *et al.*, "Prediction of allogeneic hematopoietic stem-cell transplantation mortality 100 days after transplantation using a machine learning algorithm: A European group for blood and marrow transplantation acute leukemia working party retrospective data mining stud," *J. Clin. Oncol.*, vol. 33, no. 28, pp. 3144–3151, 2015, doi: 10.1200/JCO.2014.59.1339.

[41]  R. Shouval *et al.*, "Prediction of Hematopoietic Stem Cell Transplantation Related Mortality- Lessons Learned from the In-Silico Approach: A European Society for Blood and Marrow Transplantation Acute Leukemia Working Party Data Mining Study," *PLoS One*, vol. 11, no. 3, pp. 1–14, 2016, doi: 10.1371/journal.pone.0150637.

[42]  Bone Marrow Transplant : Children Dataset https://archive.ics.uci.edu/ml/datasets/Bone+marrow+transplant%3A+children

[43]  P. Paokanta, "β-Thalassemia Knowledge Elicitation Using Data Engineering: PCA, Pearson's Chi Square and Machine Learning," *Int. J. Comput. Theory Eng.*, vol. 4, no. 5, pp. 702–706, 2012, doi: 10.7763/ijcte.2012.v4.561.

[44]  B. Amarnath and S. Appavu Alias Balamurugan, "Review on feature selection techniques and its impact for effective data classification using UCI machine learning repository dataset," *J. Eng. Sci. Technol.*, vol. 11, no. 11, pp. 1639–1646, 2016.

[45]  P. Liashchynskyi and P. Liashchynskyi, "Grid Search, Random Search, Genetic Algorithm: A Big Comparison for NAS," no. 2017, pp. 1–11.

[46]  B. Zhang *et al.*, "On the Importance of Hyperparameter Optimization for Model-based Reinforcement Learning," vol. 130, 2021, [Online]. Available: http://arxiv.org/abs/2102.13651.

[47]  L. Gao and Y. Ding, "Disease prediction via Bayesian hyperparameter optimization and ensemble learning," *BMC Res. Notes*, vol. 13, no. 1, pp. 1–6, 2020, doi: 10.1186/s13104-020-05050-0.

[48]  D. M. Belete and M. D. Huchaiah, "Grid search in hyperparameter optimization of machine learning models for prediction of HIV/AIDS test results," *Int. J. Comput. Appl.*, vol. 0, no. 0, pp. 1–12, 2021, doi: 10.1080/1206212X.2021.1974663.

[49]  S. Jolly and N. Gupta, "Understanding and Implementing Machine Learning Models with Dummy Variables with Low Variance," *Adv. Intell. Syst. Comput.*, vol. 1165, pp. 477–487, 2021, doi: 10.1007/978-981-15-5113-0_37.

[50]  J. Jo, "Effectiveness of Normalization Pre-Processing of Big Data to the Machine Learning Performance," vol. 14, no. 3, pp. 547–552, 2019.

[51]  A. Gudyś, M. Sikora, and Ł. Wróbel, "RuleKit: A comprehensive suite for rule-based learning," *Knowledge-Based Syst.*, vol. 194, p. 105480, 2020, doi: 10.1016/j.knosys.2020.105480.

[52]  Ł. Wróbel, M. Sikora, K. Kałwak, and M. Mielcarek, "Application of rule induction to discover survival factors of patients after bone marrow transplantation," *J. Med. Informatics Technol.*, vol. 22, pp. 35–53, 2013.

[53]  K. Karami, M. Akbari, M.-T. Moradi, B. Soleymani, and H. Fallahi, "Survival prognostic factors in patients with acute myeloid leukemia using machine learning techniques," *PLoS One*, vol. 16, no. 7, p. e0254976, Jul. 2021, doi: 10.1371/JOURNAL.PONE.0254976.

[54]  V. Leclerc, M. Ducher, A. Ceraulo, Y. Bertrand, and N. Bleyzac, "A Clinical Decision Support Tool to Find the Best Initial Intravenous Cyclosporine Regimen in Pediatric Hematopoietic Stem Cell Transplantation," *J. Clin. Pharmacol.*, pp. 1–19, 2021, doi: 10.1002/jcph.1924.

[55]  R. Bortnick *et al.*, "Hematopoietic stem cell transplantation in children and adolescents with GATA2-related myelodysplastic syndrome," *Bone Marrow Transplant.*, no. May, 2021, doi: 10.1038/s41409-021-01374-y.

[56]  V. Hazar *et al.*, "Risk factors predicting the survival of pediatric patients with relapsed/refractory non-Hodgkin lymphoma who underwent hematopoietic stem cell transplantation: a retrospective study from the Turkish pediatric bone marrow transplantation registry," *Leuk. Lymphoma*, vol. 59, no. 1, pp. 85–96, 2018, doi: 10.1080/10428194.2017.1330472.

[57]  Y. T. Jiaqian Qi, Tao You, Hong Wang, Sensen Shi, "Prognostic value and prediction of Bleeding in Patients undergoing Hematopoietic Stem-Cell Transplantation: A Retrospective Study of Chinese Group for Blood and Marrow Transplantation Working Party."

[58]  S. Averill, C. Ren, J. Slaven, J. Skiles, and A. Shah, "Choice of Diffusing Capacity Hemoglobin Correction Equation and Prediction of Mortality and Pulmonary Outcomes in Children Receiving Hematopoietic Stem Cell Transplantation," pp. 1–5, 2021.

[59] K. Sweiss *et al.*, "Combined immune score of lymphocyte to monocyte ratio and immunoglobulin levels predicts treatment-free survival of multiple myeloma patients after autologous stem cell transplant," *Bone Marrow Transplant.*, vol. 55, no. 1, pp. 199–206, 2020, doi: 10.1038/s41409-019-0681-3.

[60] Y. M. Liu *et al.*, "Health-related quality of life predicts length of hospital stay and survival rates for pediatric patients receiving allogeneic hematopoietic cell transplantation," *Qual. Life Res.*, 2021, doi: 10.1007/s11136-021-02887-1.

[61] S. Corbacioglu *et al.*, "Diagnosis and severity criteria for sinusoidal obstruction syndrome/veno-occlusive disease in pediatric patients: A new classification from the European society for blood and marrow transplantation," *Bone Marrow Transplant.*, vol. 53, no. 2, pp. 138–145, 2018, doi: 10.1038/bmt.2017.161.

[62] L. Stern *et al.*, "Mass Cytometry for the assessment of immune reconstitution after hematopoietic stem cell transplantation," *Front. Immunol.*, vol. 9, no. JUL, 2018, doi: 10.3389/fimmu.2018.01672.